\def\year{2020}\relax
\documentclass[letterpaper]{article} 
\usepackage{aaai20}  
\usepackage{times}  
\usepackage{helvet} 
\usepackage{courier}  
\usepackage[hyphens]{url}  
\usepackage{graphicx} 

\usepackage{amsmath,amsfonts,amssymb }
\usepackage[utf8]{inputenc}
\usepackage[english]{babel}
\usepackage[flushleft]{threeparttable}
\usepackage{hyperref}
\usepackage{multirow}
\usepackage{tabularx}
\usepackage{lettrine}
\usepackage{graphicx}
\usepackage{algorithm}
\usepackage{algorithmic}
\usepackage{subfigure}
\usepackage{amsthm}
\usepackage{color}
\usepackage{rotating}
\usepackage{stfloats}
\usepackage{flushend}
\usepackage{amsfonts}       
\usepackage{amsmath}
\newcommand{\xvec}{{\bf x}}

\usepackage{graphicx}  
\title{Dual Adversarial Co-Learning for Multi-Domain Text Classification}
\author{
Yuan Wu \and Yuhong Guo\\
School of Computer Science, Carleton University, Canada\\
yuanwu@cmail.carleton.ca, yuhong.guo@carleton.ca
}

\begin{document}

\maketitle

\begin{abstract}
With the advent of deep learning, the performance of text classification models have been improved significantly. 
Nevertheless, the successful training of a good classification model requires a sufficient amount of labeled data,
while it is always expensive and time consuming to annotate data.
With the rapid growth of digital data, similar classification tasks can typically occur in multiple domains,
while the availability of labeled data can largely vary across domains.
Some domains may have abundant labeled data, while in some other domains there may only exist 
	a limited amount (or none) of labeled data. 
Meanwhile text classification tasks are highly domain-dependent --- a text classifier trained in one domain may not perform well in another domain. 
In order to address these issues, in this paper we propose a novel dual adversarial 
	co-learning approach for multi-domain text classification (MDTC). 
The approach learns shared-private networks for feature extraction 
and deploys dual adversarial regularizations to align features across different domains
and between labeled and unlabeled data simultaneously under a discrepancy based co-learning framework,
aiming to improve the classifiers' generalization capacity with the learned features.
We conduct experiments on multi-domain sentiment classification datasets.
The results show the proposed approach achieves the state-of-the-art MDTC performance.
\end{abstract}

\section{Introduction}
Text classification is a widely existing problem in various real-world applications, 
such as spam, fraud, and sentiment analysis \cite{jindal2007review,ngai2011application,medhat2014sentiment}. 
Many similar text classification tasks can typically occur in multiple domains, while
the availability of labeled data can largely vary across domains. 
However, many text classification tasks are highly domain-dependent, 
as the same word in different domains may convey different meanings \cite{glorot2011domain,pang2008opinion}. 
For example, in the domain of sports news, the word “fast” is usually positive: “This athlete runs really fast”. 
However, “fast” is frequently used as a negative word in the domain of electronics product review; 
e.g., in the following review sentence: “the battery of this digital camera runs fast”. 
Thus, a model trained on a specific domain may fail to perform well on another domain. 
A simple solution to this problem is to train domain-specific text classifiers for each domain \cite{blitzer2007biographies}. 
Unfortunately, annotating sufficient amount of data in each domain is always expensive and time consuming, sometimes even impossible.

Multi-domain text classification (MDTC) 
\cite{li2008multi} aims to address the problems above by 
simultaneously utilizing all available data across multiple domains to improve the classification accuracy 
and improving the classifiers 
trained on each specific domain with the help of other related domains. 
Recently, deep neural network based models have been used for multi-domain learning problem in 
both computer vision \cite{mansour2009domain,duan2012domain} and natural language processing fields \cite{wu2015collaborative,chen2018multinomial}.
Some previous work tackles MDTC 
by training a shared feature extractor across different domains 
for useful information sharing \cite{liu2016recurrent}. 
However, focusing only on a shared feature subspace across different domains may ignore the domain-dependent characteristics of each domain.  
One way to address this problem is to learn both shared feature extractor and domain-specific feature extractor  
through a shared-private scheme \cite{bousmalis2016domain}.
However, the traditional shared-private models suffer from problems such 
like domain-dependant features creep into the shared latent subspace \cite{liu2017adversarial}, 
which consequently can result in performance degradation. 
Moreover with the reality of lacking sufficient labeled data, 
the features extracted in such models may overfit the labeled training data
and fail to generalize well.

To address these problems, in this paper we propose a novel dual adversarial co-learning approach for MDTC. 
The approach extracts both domain-invariant and domain-specific features through shared-private networks 
and learns two classifiers on the extracted features.
The classifiers and feature extractors are 
co-learned in an adversarial manner based on the prediction discrepancy on unlabeled data.  
Meanwhile, a multinomial multi-domain adversarial discriminator 
is deployed to enhance the effective extraction of domain-invariant features,
while separating them from the domain-specific features.
Different from previous existing multi-domain learning methods, 
the proposed approach not only attempts to align data across domains in the extracted feature space
but also simultaneously align labeled and unlabeled data in each domain, avoiding overfitting to the limited labeled data.
The overall learning is conducted through three adversarial steps,
aiming to produce feature representations that not only generalize well
across domains but also generalize into the unlabeled data.
We evaluate the proposed approach's efficacy by conducting experiments on two popular MDTC data sets,
and also extend the approach for unsupervised domain adaptation. 
The experimental results show the proposed approach achieves the state-of-the-art performance in both scenarios.

\section{Related Work}

\subsection{Multi-Domain Text Classification}

Text classification such as review classification is one important problem 
with wide real application needs in many different domains \cite{pang2002thumbs}
and it is also well known as a highly domain dependent problem. 
With the varying availability of labeled data in different but similar domains,
multi-domain text classification has been studied to improve text classification performance
by exploiting data in multiple domains together \cite{li2008multi}.
With multi-domain classification, classifiers can be trained for domains with scarce labels
by utilizing other related domains with abundant labeled data. 
Some early works have adopted transfer learning techniques for MDTC. 
The structural correspondence learning (SCL) algorithm \cite{blitzer2007biographies} computes the association between pivot features among different domains in order to capture correspondences among features. 
\cite{pan2010cross} proposed a spectral feature alignment (SFA) algorithm for MDTC by reducing the gap between different domains. The most recent prior works on MDTC include Collaborative Multi-Domain Sentiment Classification (CMSC) \cite{wu2015collaborative}, Recurrent Neural Network for Text Classification with Multi-Task Learning (RNN-MT) \cite{liu2016recurrent}, 
Multinomial Adversarial Networks for Multi-Domain Text Classification (MAN) \cite{chen2018multinomial} and Adversarial Multi-task Learning for Text Classification (ASP-MTL) \cite{liu2017adversarial}. 
CMSC uses two types of classifiers, a classifier shared by all the domains and a set of classifiers developed for each domain, and it requires external resources to help improve classification accuracy. 
RNN-MT used recurrent neural networks to map data from different domains into a common latent space, then used the shared features to train the model. Both MAN and ASP-MTL leverage standard adversarial training, 
as well as the shared-private model \cite{bousmalis2016domain}, to guide the feature extraction to guarantee that the shared feature extractor only generates common and domain-invariant features. 
Our proposed work further advances the line of study by deploying dual adversarial co-training.
We assume that the amount of labeled data in each domain is varying and insufficient, aiming to train accurate classifiers by leveraging all the available resources across domains.

\subsection{Adversarial Training}

The idea of adversarial training was initialized by the generative adversarial network (GAN) on image generation \cite{goodfellow2014generative}.
The GAN model 
formulates the generator training problem as a minimax adversarial game
and deploys a discriminator, which discriminates the generated fake samples from the real samples, 
as an adversarial player,
aiming to learn a robust generator that can fool the discriminator.
The adversarial learning mechanism later is also exploited for domain adaptation
\cite{ganin2016domain,bousmalis2016domain}, 
where the discriminator tries to differ data in the source domain from that in the target domain,
aiming to help learn domain-invariant features. 
The work in \cite{saito2018maximum} deploys adversarial learning 
to align the distributions of source and target domains by adjusting the task-specific decision boundaries. 
It plays a min-max game between two classifiers and one feature extractor,
and uses the prediction disagreement of the two classifiers in the target domain as a discrimination.
Nevertheless, most prior works applied adversarial training in a single-source and single-target scenario. 
The recent work \cite{zhao2018adversarial} extends adversarial training to multiple-source scenarios for domain adaptation,
which uses a domain classifier as the adversarial discriminator.
Our proposed method uses a novel dual adversarial training for MDTC tasks, 
which learns feature extractors that not only align multiple domains, but also align labeled and unlabeled data.

\begin{figure*}
    \centering
    \includegraphics[width=5.2in,height=1.9in]{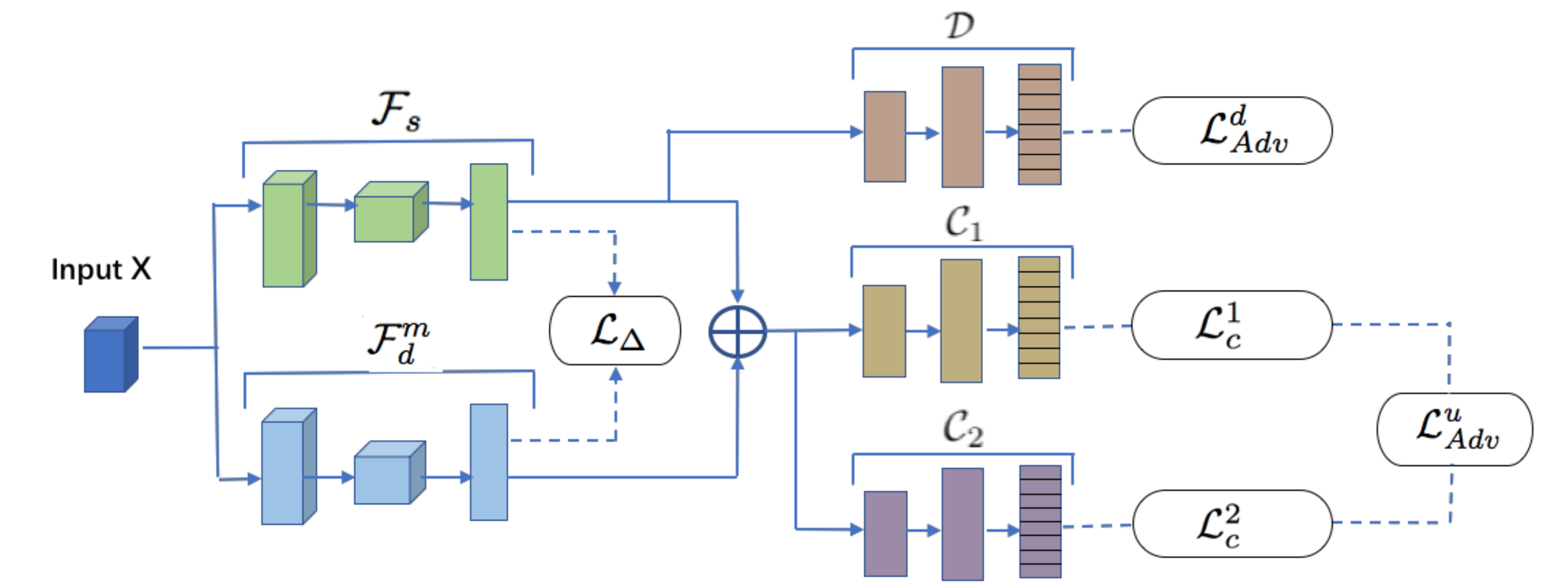}
    \vskip -.05in
    \caption{Architecture of the dual adversarial co-learning model. 
	A shared feature extractor $\mathcal{F}_s$ learns to capture domain-invariant features. 
	Each domain-specific feature extractor $\mathcal{F}^m_{d}$ 
	learns to capture domain-dependent features. 
	A regularizer $\mathcal{L}_{\Delta}$ is used to separate the shared and domain-dependent features. 
	A domain discriminator $\mathcal{D}$ is used to adversarially help identify domain-invariant 
	features through an adversarial loss $\mathcal{L}^d_{adv}$. 
	Two classifiers $\mathcal{C}_1$ and $\mathcal{C}_2$ are co-learned to adversarially enforce the alignment of labeled and unlabeled data in each domain in the extracted feature space.
	}
    \label{fig:my_label_1}
    \vskip -.1in
\end{figure*}

\section{Approach}

In this work, we consider MDTC tasks in the following setting. Assume we have $M$ domains, 
each $m$-th domain has a limited number of labeled instances $\mathbb{L}_m=\{(\xvec_i,y_i)\}_{i=1}^{l_m}$ and 
a set of unlabeled instances $\mathbb{U}_m=\{\xvec_i\}_{i=1}^{u_m}$. 
The problem of MDTC is to utilize all the available resources across the $M$ domains to 
improve the overall multi-domain classification performance. 
In this section, 
we present a novel dual adversarial co-learning model and 
a stepwise adversarial training algorithm for MDTC.

\subsection{Dual Adversarial Co-Learning Model}

As each domain contains limited labeled data, the main idea of MDTC is to exploit the joint labeled resources in multiple domains
to capture common generalizable information.
Towards this goal, we propose a novel dual adversarial co-learning model for MDTC,
which is illustrated in Figure \ref{fig:my_label_1}.
The model has five component networks: a shared feature extraction network $\mathcal{F}_s$, 
a set of domain-specific feature extraction networks $\{\mathcal{F}^m_{d}\}_{m=1}^M$, 
two classification networks $\mathcal{C}_1$ and $\mathcal{C}_2$ that use features from both types of feature extractors, 
and a domain discriminator $\mathcal{D}$. 
The shared feature extractor $\mathcal{F}_s$ learns to capture the shared features that can contribute to the classification tasks across all domains, 
while each domain-specific feature extractors $\mathcal{F}^m_{d}$ 
learns to capture domain-dependent features that contribute specifically only to their own domain. 
These feature extractors perform data representation learning
and can adopt the form of Convolutional Neural Network (CNN), Recurrent Neural Network (RNN), 
or Multiple Layer Perceptron (MLP), depending on the particular task. 
Each instance in the training data will go through both the shared feature extraction network
and its domain-specific feature extraction network 
to produce two feature vectors for the consequent classification task.
The proposed model deploys two adversarial learning mechanisms to induce 
useful and generalizable features that not only align data across the multiple domains but also 
align the labeled data and unlabeled data in each domain
through co-learning of a pair of classifiers.
We introduce each of the adversarial learning mechanisms below.

\subsubsection{Domain Discrimination based Adversarial Learning}

In the proposed model, the shared feature extractor $\mathcal{F}_s$  
and the domain-specific feature extractors $\{\mathcal{F}^m_{d}\}_{m=1}^M$ 
are expected to complement each other and maximumly capture useful data representations
for multiple domains. 
The shared features extracted are the domain generalizable features that enable
knowledge sharing across domains. 
It is important to capture the right domain-invariant features and 
prevent the shared and domain-specific features from interfering with each other.
We propose to deploy a multinomial domain discriminator $\mathcal{D}$ to discriminate data from different domains
in the learned domain-invariant feature space. For $M$ domains, $\mathcal{D}$ 
will be a $M$-class classifier and output a probability vector on each input instance;
e.g., $\mathcal{D}_m(\mathcal{F}_s(\xvec))$ denotes the predicted probability of instance $\xvec$ coming from the $m$-th domain. 
Intuitively, the multinomial discriminator $\mathcal{D}$ should be optimized to maximumly discriminate instances from different domains,
while the shared feature extractor $\mathcal{F}_s$ should be optimized to maximumly fool the discriminator. 
If a strong discriminator can not identify the domains of input instances with the learned features, 
these features are essentially domain-invariant. 
We encode this intuition through the following adversarial learning formulation:
\begin{align}
	\min_{\mathcal{F}_s}\max_{\mathcal{D}} \quad
	\mathcal{L}^d_{Adv}= \sum_{m=1}^M \mathbb{E}_{\xvec\sim\mathbb{L}_m\cup\mathbb{U}_m} 
	{\log[ \mathcal{D}_m(\mathcal{F}_s(\xvec))]}
	\label{eq:advD}
\end{align}
This multinomial discriminator-based formulation 
extends the standard binary adversarial discriminator into the multi-domain learning scenarios
under the same adversarial learning principle.

\subsubsection{Separation regularizer}
Motivated by some recent works \cite{bousmalis2016domain,liu2017adversarial} on network separation analysis, 
we also introduce a separation regularizer to ensure the difference of the domain-specific extractors from the domain-invariant feature
extractor. 
Specifically, the separation regularizer minimizes the similarity between the extracted domain-invariant features
and the domain-specific features on the same data in all the domains:
\begin{align}
	\min_{\mathcal{F}_s,\{\mathcal{F}_d^m\}} \mathcal{L}_{\Delta}= 
	\sum_{m=1}^{M}\Big\|\sum_{\xvec\sim\mathbb{L}_m} \mathcal{F}_s(\xvec) \mathcal{F}^m_d(\xvec)^\top \Big\|_F^2
\end{align}
where $\mathcal{F}_s(\xvec)$ and $\mathcal{F}^m_d(\xvec)$
denote the column feature vectors extracted,
and ${\| \cdot \|}_F$ denotes the Frobenius norm. 
By minimizing such a separation regularizer, the difference between the shared feature extractor and each domain-specific feature
extractor can be enforced.

\subsubsection{Prediction Discrepancy based Adversarial Learning}

After feature extraction,
two classification networks $\mathcal{C}_1$ and $\mathcal{C}_2$ 
are deployed to take the concatenation of shared and domain-specific features as input
and predict the class labels of the given instance.
The networks $\mathcal{C}_1$ and $\mathcal{C}_2$ can be MLPs with a softmax output layer
that produces prediction probabilities in each class.
Both classifiers can be trained together with the feature extractors
by minimizing the prediction loss on the labeled instances:
\begin{align*}
	&\min_{\mathcal{C}_1, \mathcal{C}_2, \mathcal{F}_s,\{\mathcal{F}_d^m\}} 
	\quad \mathcal{L}_c^1 + \mathcal{L}_c^2, 
	\\
	\mbox{where} &
	\quad \mathcal{L}_c^i = 
	-{\sum_{m=1}^M \mathbb{E}_{(\xvec,y)\sim{\mathbb{L}_m}}} 
	\log[\mathcal{C}_{iy}(\mathcal{F}_s(\xvec),\mathcal{F}^m_{d}(\xvec))],
\end{align*}
$i\in\{1,2\}$ denotes the index of two classifiers,
and $\mathcal{C}_{iy}(\mathcal{F}_s(\xvec),\mathcal{F}^m_{d}(\xvec))$
denotes the prediction probability of instance $\xvec$ belonging to the $y$-th class
by the classifier $\mathcal{C}_i$.
Negative loglikelihood is used as the loss function.
Then $\mathcal{L}_c^1$ and $\mathcal{L}_c^2$ denote the prediction loss
of the two classifiers on the labeled data in $M$ domains respectively.

As we have limited labeled data in each domain, to avoid overfitting of the features
to the labeled data and ensure their generalizability to unlabeled instances,
we adopt another adversarial learning mechanism based on the prediction discrepancy
of the two classifiers on the unlabeled data to assist feature representation learning,
especially the domain-specific feature learning:
\begin{align}
	& \min_{\mathcal{F}_s,\{\mathcal{F}_d^m\}}
	\max_{\mathcal{C}_1,\mathcal{C}_2:d(\mathcal{C}_i,\mathcal{C}^*_i)\leq\epsilon}\quad
	\mathcal{L}_{Adv}^u \\
	\mbox{where}\quad &
	\mathcal{L}_{Adv}^u= \sum_{m=1}^{M} \mathbb{E}_{\xvec\sim\mathbb{U}_m} 
	{\left\|
	\begin{array}{l}
		\mathcal{C}_{1}(\mathcal{F}_s(\xvec),\mathcal{F}^m_{d}(\xvec)) -\\[.5ex]
		\mathcal{C}_{2}(\mathcal{F}_s(\xvec),\mathcal{F}^m_{d}(\xvec)) 
	\end{array}
	\right\|_1}
	\label{eq:advU}
\end{align}
where $\|\cdot\|_1$ denotes the $\ell_1$ norm function
and we use $\ell_1$ norm to measure the prediction difference between the two classifiers on the unlabeled data,
$d(\cdot,\cdot)$ denotes a distance function,
and $\mathcal{C}^*_{i}$ ($i=1,2$) denotes the classifiers trained on the labeled data.
The constraint $d(\mathcal{C}_i,\mathcal{C}^*_i)\leq\epsilon$
enforces $\mathcal{C}_1$ and $\mathcal{C}_2$ to be in the close neighborhood 
of the original classifiers $\mathcal{C}^*_1$ and $\mathcal{C}^*_2$.
This adversarial mechanism can induce features that are robust to small changes in the classifiers.
By maximizing the prediction discrepancy of the two classifiers, $\mathcal{C}_1$ and $\mathcal{C}_2$, on unlabeled data 
in the close neighborhood of the original classifiers (within distance $\epsilon$), 
unlabeled instances that are out of the scope of or not aligned with the labeled instances 
in the joint feature representation space can be identified.
These instances are most likely located near the classification boundary 
of the original classifiers ($\mathcal{C}^*_1, \mathcal{C}^*_2$) 
-- thus some near neighbor classifiers ($\mathcal{C}_1, \mathcal{C}_2$) can easily disagree on them.
With such an adversarial opponent, by 
minimizing the prediction agreement on unlabeled data, 
the feature generators can be tuned to produce feature representations
in which the unlabeled instances can be well aligned with the labeled instances
and hence the classifiers can have better generalization capacities.
%

\subsection{Stepwise Adversarial Training}

To integrate the two adversarial mechanisms introduced above
into the learning process of our feature extractors and classifiers,
we develop a stepwise adversarial training procedure. 
The procedure has three sequential steps, model learning step, adversarial opponent step,
and model refinement step, which work together to enforce robust domain adaptation.
We elaborate the three steps below.

\subsubsection{Model Learning Step}
The proposed MDTC model mainly comprises of the feature extraction networks, 
$\mathcal{F}_s$ and $\{\mathcal{F}^m_{d}\}$, 
and the classification networks, $\mathcal{C}_1$ and $\mathcal{C}_2$. 
We first conduct initial classifier training over them by minimizing 
the separation regularized prediction loss on the labeled training data in multiple domains:
\begin{align}
	\min_{\mathcal{F}_s,\{\mathcal{F}^m_{d}\},\mathcal{C}_1,\mathcal{C}_2} 
	\mathcal{L}^1_{c}+\mathcal{L}^2_{c} + \alpha \mathcal{L}_{\Delta}
\end{align}
where $\alpha$ is a trade-off parameter.
The separation regularizer is used here to avoid negative interference between the domain invariant and domain specific features.
This step performs standard training on the labeled instances.
The model produced needs to be further refined through adversarial learning
to improve the feature extractors.

\subsubsection{Adversarial Opponent Step}
We then take the two adversarial mechanisms in Eq.(\ref{eq:advD}) and Eq.(\ref{eq:advU}) into consideration.
Instead of performing an alternating minimax optimization, 
we simply conduct two discrete steps: 
maximization over the opponents (discriminator and the classifiers) 
in this step and minimization over the model (feature extractors) in the next step.
To avoid using constraints to limit the search for adversarial classifiers within a close neighborhood
of the initial classifiers, we reformulate the adversarial objective by 
taking the prediction loss on the labeled data into account:
\begin{align}
	\max_{\mathcal{C}_1, \mathcal{C}_2 }\quad &
	-\left(\mathcal{L}^1_{c}+\mathcal{L}^2_{c}\right) + \mathcal{L}^u_{Adv} 
	\\
	\max_{\mathcal{D}}\quad &
	\mathcal{L}^d_{Adv} 
\end{align}
These adversarial maximizations identify the difficulties 
in aligning domains in the shared feature space and aligning labeled and unlabeled data in the joint feature space
respectively,
which provide directions for further refining the feature extractors in the next step.

\subsubsection{Model Refinement Step}
In this last step, 
we refine the feature extractors ($\mathcal{F}_s$ and $\{\mathcal{F}^m_{d}\}$) 
to fool the discriminator and minimize the prediction discrepancy on unlabeled data:
\begin{align}
	\min_{\mathcal{F}_s, \{\mathcal{F}^m_{d}\}}\quad  \mathcal{L}^u_{Adv}+\gamma \mathcal{L}^d_{Adv}	
\end{align}
The refined features are expected to be more robust and generalizable for better classifier training.

We adopt a mini-batch based stochastic gradient descent algorithm for this 
stepwise adversarial training, which is illustrated in Algorithm 1.
In each iteration of the algorithm, 
the three steps, L-step  
(model Learning step), A-step
({A}dversarial opponent step),
and R-step (model {R}efinement step),
can be conducted sequentially 
over the sampled batches,
while 
each step performs a stochastic gradient based update regarding the
related objective. 

\begin{algorithm}
\caption{Stochastic gradient descent training algorithm}\label{trainingalg}
\begin{algorithmic}[1]
\STATE{\bf Input:} labeled data $\mathbb{L}_m$ and unlabeled data $\mathbb{U}_m$ in $M$ domains; 
	hyper-parameters $\alpha$, $\gamma$.
\FOR{number of training iterations}
	\STATE Sample labeled mini-batches from the multiple domains $B^\ell=\{B^\ell_1,\cdots, B^\ell_M\}$.
	\STATE Sample unlabeled mini-batches from the multiple domains $B^u=\{B^u_1,\cdots, B^u_M\}$.
	\STATE \underline{L-Step}: 
	Calculate $\mathcal{L}_L=\mathcal{L}^1_c +\mathcal{L}^2_c+\alpha \mathcal{L}_\Delta$ on $B^\ell$;\\ 
	\hspace{.37in} Update $\mathcal{F}_s$, $\{\mathcal{F}^m_{d}\}$, $\mathcal{C}_1$, $\mathcal{C}_2$ 
	by descending along\\\hspace{.4in} gradients $\Delta{\mathcal{L}_L}$.\\[.2ex]
	\STATE \underline{A-Step}: Calculate $\mathcal{L}_A=-(\mathcal{L}^1_c +\mathcal{L}^2_c)+ \mathcal{L}^u_{Adv}$ 
	on $B^\ell$\\\hspace{.4in} and $B^u$, calculate $\mathcal{L}^d_{Adv}$ on $B^\ell$ and $B^u$;\\ 
	\hspace{.4in} Update $\mathcal{C}_1$ and $\mathcal{C}_2$ by ascending along $\Delta{\mathcal{L}_A}$;\\
	\hspace{.4in} Update $\mathcal{D}$ by ascending along $\Delta{\mathcal{L}^d_{Adv}}$.\\[.2ex]
	\STATE \underline{R-Step}: Calculate $\mathcal{L}_R= \mathcal{L}^u_{Adv} +\gamma\mathcal{L}^d_{Adv}$;\\ 
	\hspace{.4in} Update $\mathcal{F}_s$, $\{\mathcal{F}^m_{d}\}$ by descending along $\Delta{\mathcal{L}_R}$.\\[.2ex]
\ENDFOR
\end{algorithmic}
\end{algorithm}

\section{Experiments}
We conduct experiments on two multi-domain text classification datasets, and also extend the empirical study
into the scenario of multi-source unsupervised domain adaptation.
In this section we report the experimental setting and results.

\subsection{Experimental Settings}
\paragraph{Dataset}
We conducted experiments 
on two multi-domain text classification (MDTC) datasets: the multi-domain Amazon review dataset \cite{blitzer2007biographies} 
and the FDU-MTL dataset \cite{liu2017adversarial}. 
Both datasets are widely used in multi-domain and cross domain text classification tasks. 
The Amazon review dataset 
has four domains, i.e., book, DVD, electronics and kitchen. Each domain contains 2,000 samples: 1,000 positive reviews and 1,000 negative reviews. 
Note that all reviews in the dataset were already encoded into 5,000 dimensional vectors of bag-of-word unigram and bigram features
with binary labels, which have lost all word order information
and can not be processed by CNNs or RNNs anymore.
We hence used MLPs as our feature extractors, which have an input size of 5,000. 
More specifically, for the MDTC experiment on the Amazon review dataset, we followed the experiment setting in \cite{wu2015collaborative} and conducted a 5-fold cross validation test. We randomly divide data in each domain into five partitions with equal size, where three partitions are used for training, one serves as the validation set, and the remaining one is used for testing. 
The 5-fold average test classification accuracy is recorded.

As the Amazon dataset has fewer domains,
the FDU-MTL dataset 
consists of 16 domains, each domain 
contains reviews with binary sentiment labels. The first 14 domains are Amazon product reviews on different products such as books, electronics, DVD, kitchen, apparel, camera, health, music, toys, video, baby, magazine, software and sports. 
The remaining two domains are movie reviews from the IMDb and MR datasets. 
All the data are collected as raw text data only being tokenized by the Stanford tokenizer. 
Hence this dataset allows one to use more advanced feature extractors. 
The data in each domain is randomly divided into three parts: training set (70$\%$), validation set (10$\%$) and testing set (20$\%$). There exists 200 samples in the validation set and 400 samples in the testing set for each domain. While the numbers of labeled and unlabeled instances vary across domains they are roughly 1400 and 2000, respectively.

\paragraph{Implementation Details}
The proposed model has two hyperparameters, $\alpha$ and $\gamma$. In the experiments, we set $\alpha=0.1$ and $\gamma=0.1$.
We set the extracted shared feature dimension as 128 and the extracted domain-specific feature dimension as 64.
$\mathcal{C}_1$, $\mathcal{C}_2$ and $\mathcal{D}$ are MLPs with one hidden layer with 
128, 64 and 128 hidden units resepctively.
ReLU is used as the activation function. 
For the experiments on Amazon review dataset, we use MLPs with two hidden layers (with 1000 and 500 units respectively)
as feature extractors. 
The input size of the MLPs is set to 5000. 
For the experiments on FDU-MTL, we use CNN with one convolutional layer as our feature extractor. 
It uses different kernel sizes (3, 4, 5), and the number of kernels are 200. 
The input of the convolutional layer is the 100-dim word embeddings, obtained using word2vec \cite{mikolov2013efficient}, 
of each word in the input sequence. 
We use Adam optimizer \cite{kingma2014adam} to train our models, with the learning rate 0.0001. 
We use a batch size of 8, and 50 training iterations.
On the test data, we use the average prediction probability scores of the two classifiers
to determine the prediction labels.

\begin{table*}[t]
\caption{\label{font-table} MDTC classification accuracies on the Amazon review dataset. 
	Bold font denotes the best classification results.}
\vskip .1in	
\label{table_ref1}
\centering
\setlength{\tabcolsep}{4pt}
\begin{tabular}{ l  c c c c c c c c}
\hline
	Domain & MTLGraph  & CMSC-LS & CMSC-SVM & CMSC-Log & MAN-L2 & MAN-NLL & DACL(Proposed)\\
\hline
Books &  79.69 &  82.10 & 82.26 & 81.81 & 82.46 & 82.98 & $\mathbf{83.45}$ \\
DVD &  81.84 &  82.40 & 83.48 & 83.73 & 83.98 & 84.03 & $\mathbf{85.50}$ \\
Elec.  & 83.69  & 86.12 & 86.76 & 86.67 & 87.22 & 87.06 & $\mathbf{87.40}$ \\
Kit.  & 87.06 &  87.56 & 88.20 & 88.23 & 88.53 & 88.57 & $\mathbf{90.00}$\\
\hline
AVG  & 83.06 &  84.55 & 85.18 & 85.11 & 85.55 & 85.66 & $\mathbf{86.59}$\\
\hline
\end{tabular}
\end{table*}
\begin{table*}[t]
\caption{\label{font-table} MDTC classification accuracies on the FDU-MTL dataset. 
	Bold font denotes the best classification accuracy results.}
\vskip .1in	
\label{table_ref2}
\centering
\begin{tabular}{ l c c c c c c}
\hline
	Domain & MT-CNN & MT-DNN & ASP-MTL & MAN-L2 & MAN-NLL & DACL(Proposed)\\
\hline
books & 84.5 & 82.2 & 84.0 & $\mathbf{87.6}$ & 86.8 & 87.5 \\
electronics & 83.2 & 81.7 & 86.8 & 87.4 & 88.8 & $\mathbf{90.3}$ \\
dvd & 84.0 & 84.2 & 85.5 & 88.1 & 88.6 & $\mathbf{89.8}$ \\
kitchen & 83.2 & 80.7 & 86.2 & 89.8 & 89.9 & $\mathbf{91.5}$\\
apparel & 83.7 & 85.0 & 87.0 & 87.6 & 87.6 & $\mathbf{89.5}$\\
camera & 86.0 & 86.2 & 89.2 & 91.4 & 90.7 & $\mathbf{91.5}$\\
health & 87.2 & 85.7 & 88.2 & 89.8 & 89.4 & $\mathbf{90.5}$ \\
music & 83.7 & 84.7 & 82.5 & 85.9 & 85.5 & $\mathbf{86.3}$ \\
toys & 89.2 & 87.7 & 88.0 & 90.0 & 90.4 & $\mathbf{91.3}$ \\
video & 81.5 & 85.0 & 84.5 & 89.5 & $\mathbf{89.6}$ & 88.5\\
baby & 87.7 & 88.0 & 88.2 & 90.0 & 90.2 & $\mathbf{92.0}$ \\
magazine & 87.7 & 89.5 & 92.2 & 92.5 & 92.9 & $\mathbf{93.8}$ \\ 
software & 86.5 & 85.7 & 87.2 & 90.4 & $\mathbf{90.9}$ & 90.5 \\
sports & 84.0 & 83.2 & 85.7 & 89.0 & 89.0 & $\mathbf{89.3}$ \\
IMDb & 86.2 & 83.2 & 85.5 & 86.6 & 87.0 & $\mathbf{87.3}$\\
MR & 74.5 & 75.5 & $\mathbf{76.7}$ & 76.1 & $\mathbf{76.7}$ & 76.0\\
\hline
AVG & 84.5 & 84.3 & 86.1 & 88.2 & 88.4 & $\mathbf{89.1}$ \\
\hline
\end{tabular}
\end{table*}

\begin{table}
	\setlength{\tabcolsep}{2.5pt}	
\caption{\label{font-table} Ablation study analysis.}
\vskip .1in
\label{table_ref4}
\centering
\begin{tabular}{ l c c c c c }
\hline
Method & Books & DVD & Electronics & Kitchen & AVG\\
\hline
	DACL (full)& 83.45 & 85.50 & 87.40 & 90.00 & 86.59 \\
DACL w/o $\mathcal{D}$ & 82.90 & 85.10 & 86.30 & 89.15 & 85.86 \\
DACL w/o $\mathcal{C}_2$ & 81.85 & 83.10 & 85.75 & 89.10 & 84.95 \\
\hline
\end{tabular}
\end{table}

\begin{table*}
\caption{\label{font-table} Unsupervised domain adaptation results on the Amazon review dataset. Bold numbers indicate the best classification accuracy in each domain.}
\vskip .1in
\label{table_ref3}
\centering
\setlength{\tabcolsep}{5.5pt}
\begin{tabular}{ l c c c c c c c c}
\hline
Domain & MLP & mSDA & DANN & MDAN(H) & MDAN(S) & MAN-L2 & MAN-NLL & DACL(Proposed)\\
\hline
Books & 76.55 & 76.98 & 77.89 & 78.45 & 78.63 & 78.45 & 77.78 & $\mathbf{80.22}$\\
DVD & 75.88 & 78.61 & 78.86 & 77.97 & 80.65 & 81.57 & 82.74 & $\mathbf{82.96}$\\
Elec. & 84.60 & 81.98 & 84.91 & 84.83 & $\mathbf{85.34}$ & 83.37 & 83.75 & 84.90\\
Kit. & 85.45 & 84.26 & 86.39 & 85.80 & 86.26 & 85.57 & 86.41 & $\mathbf{86.75}$\\
\hline
AVG & 80.46 & 80.46 & 82.01 & 81.76 & 82.72 & 82.24 & 82.67 & $\mathbf{83.71}$\\
\hline
\end{tabular}
\end{table*}

\subsection{Multi-Domain Text Classification}

\paragraph{Comparison Methods}

We compared our proposed dual adversarial co-learning (DACL) model with a number of state-of-the-art MDTC methods,
which are listed as follows:

\begin{itemize}
    \item MTLGraph: The Multi-task learning with graph regularization \cite{zhou2011malsar}. This method uses the sentiment word distribution similarity graph as its domain similarity graph.
    
    \item CMSC-LS, CMSC-SVM, CMSC-Log: The collaborative multi-domain sentiment classification methods with squared loss, hinge loss and log loss respectively \cite{wu2015collaborative}. 
    
    \item MAN: The multinomial adversarial network for multi-domain text classification \cite{chen2018multinomial}. This method uses two forms of loss to train the domain discriminator: The negative log-likelihood loss (MAN-NLL) and the least square loss (MAN-L2).
    
    \item MT-CNN: 
	    The deep neural network model which can learn features across different tasks given very limited prior knowledge by using a single convolutional layer \cite{collobert2008unified}.
    
    \item MT-DNN: The mult-task deep neural network model with bag-of-words input and MLPs \cite{liu2015representation}, in which a hidden layer is shared.
    
    \item ASP-MTL: The adversarial multi-task learning model for text classification \cite{li2008multi}, in which shared-private scheme and adversarial training are used to guide feature extration.
\end{itemize}

All the comparison methods use the standard partitions of the datasets. We hence take convenience to cite the results from 
\cite{wu2015collaborative,li2008multi,chen2018multinomial} for fair comparisons.

\paragraph{Experimental Result Analysis}

The experimental results on Amazon review dataset and FDU-MTL dataset are reported 
in Table~\ref{table_ref1} and Table~\ref{table_ref2} respectively.
The recent studies that used the Amazon dataset have showed it is very difficult to 
achieve multi-domain performance gains on this dataset.
From Table~\ref{table_ref1} 
we can see that by adopting the dual adversarial co-learning, 
the proposed DACL method achieves the best performance in all domains on the Amazon review dataset
comparing with the other state-of-the-art methods,
including the ones that also adopted domain adversarial training.
Moreover, the improvement gains yielded by our proposed approach 
are more notable than the performance difference between any two best comparison methods.
This suggests that the dual adversarial mechanism in our proposed approach is more effective. 

For the experimental results on FDU-MTL, reported in Table~\ref{table_ref2}, 
the proposed DACL method outperforms MT-CNN and MT-DNN consistently across all domains
with notable large performance gains. 
When compared with the state-of-the-art MAN-L2, MAN-NLL and ASP-MTL, DACL achieves the best performance on 12 out of 16 domains, 
and obtains superior result in terms of average classification accuracy. 
The experimental results again validate the efficacy of our proposed method.

\paragraph{Ablation Study}

Since the proposed DACL model has five component networks, which are involved in two adversarial mechanisms. 
We investigated how different components, importantly the adversarial mechanisms, 
in our model can impact the performance on the Amazon review dataset. 
In particular we investigated two ablation variants:
(1) DACL w/o $\mathcal{D}$, the variant of the proposed DACL model without the discriminator $\mathcal{D}$, 
which hence drops the domain-based adversarial learning;
(2) DACL w/o classifier $\mathcal{C}_2$, the variant of the proposed DACL model without the second classifier,
which hence drops the discrepancy-based adversarial learning.
The comparison results bewteen these two variants and the full model are reported in
Table~\ref{table_ref4}. 
We can see both variants induced inferior results,
and the full model with dual adversarial mechanisms produced the best results, comparing to the two ablation variants.
This validates the contribution of both adversarial mechanisms.

\begin{figure}[htbp]
\centering
\subfigure[$\alpha$]{
\includegraphics[width=5.0cm,height=3.3cm]{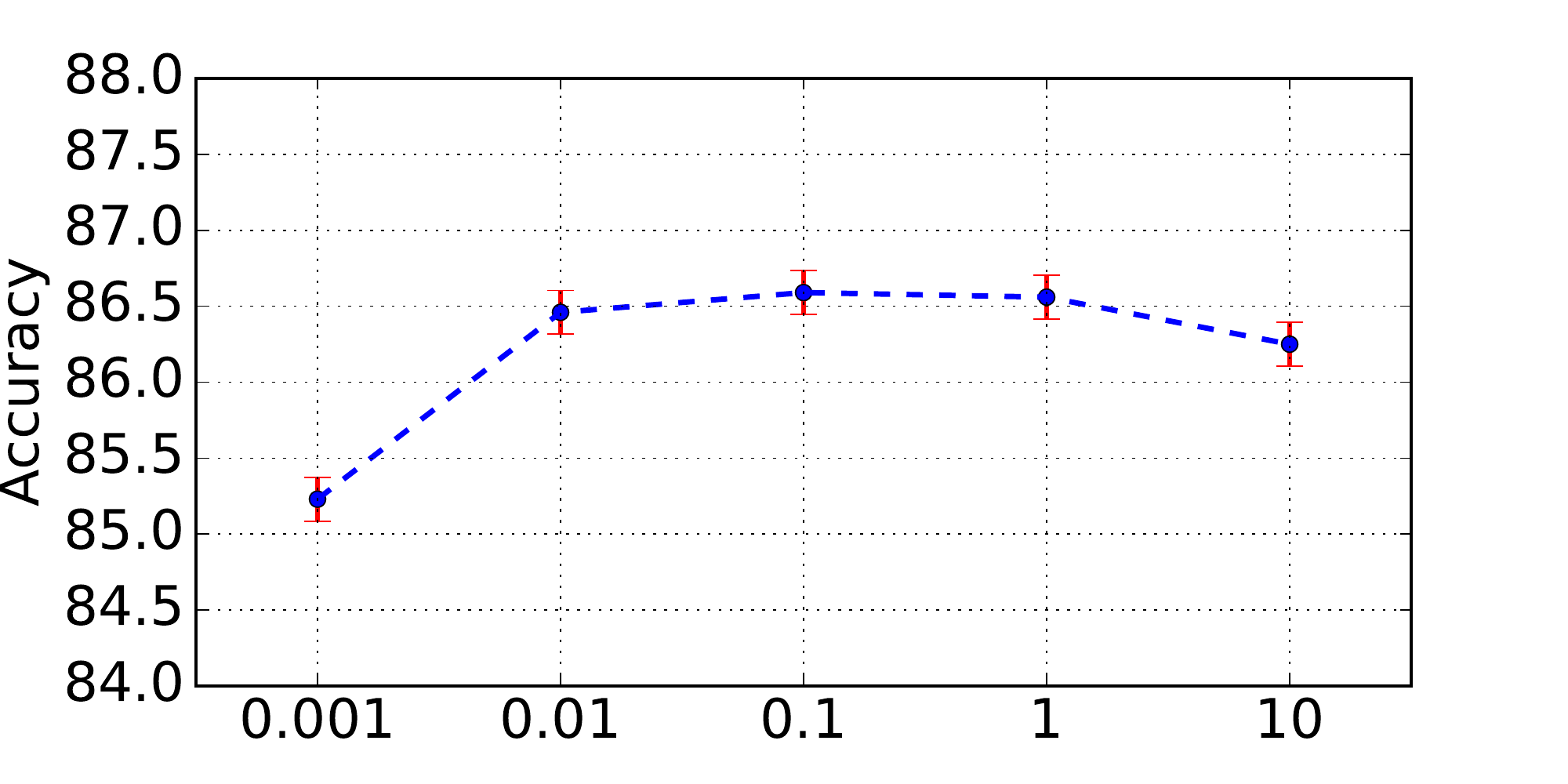}
}
\quad
\subfigure[$\gamma$]{
\includegraphics[width=5.0cm,height=3.3cm]{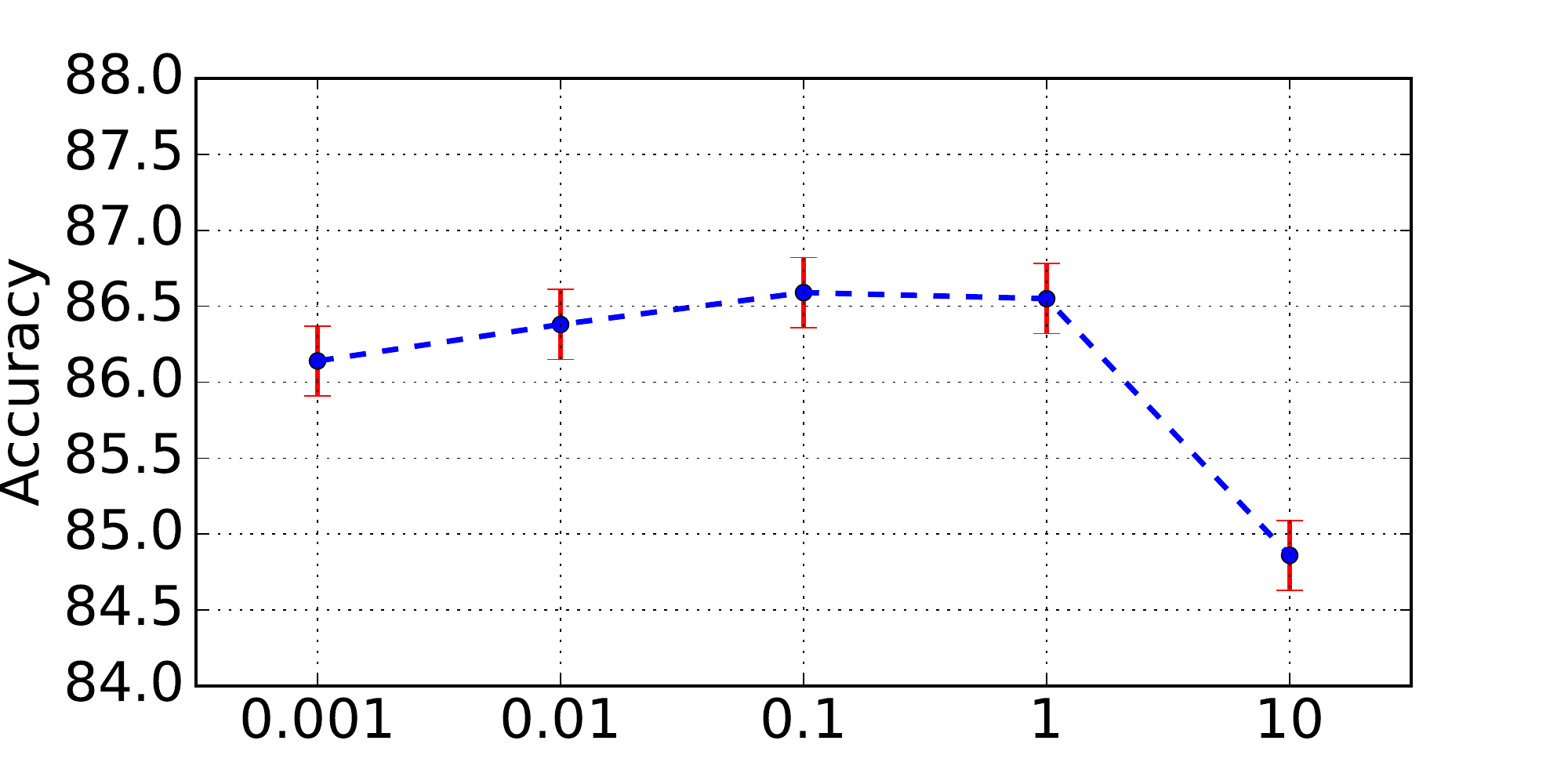}
}
\caption{Parameter sensitivity analysis}
\label{fig:my_label_2}
\end{figure}

\paragraph{Parameter Sensitivity Analysis}

The proposed model has two hyperparameters, $\alpha$ and $\gamma$. 
$\alpha$ is used to weight the feature extractor separation regularizer; 
$\gamma$ are used to balance the two adversarial losses. 
We conducted experiments on the Amazon review dataset to perform sensitivity analysis on these parameters.
We first fixed $\gamma=0.1$ and conducted experiments with different $\alpha$ values from $\{0.001, 0.01, 0.1, 1, 10\}$. 
Then we fixed $\alpha=0.1$ and conducted experiments with different $\gamma$ values from the same range $\{0.001, 0.01, 0.1, 1, 10\}$. 
The experimental results are reported in Figure \ref{fig:my_label_2}. 
The results are the average classification accuracy across multi-domains on the Amazon review dataset.
From Figure \ref{fig:my_label_2}, we can see that the average classification accuracy increases quickly 
with $\alpha$ value increasing from $0.001$ to $0.1$. 
This suggests the feature extractor separation regularizer is useful.
With $\alpha$ changing from $0.1$ to $1$, the performance change is very small,
while the further increase of $\alpha$ degrades the performance. 
This suggests this regularizer can not be over emphasized, 
and a value in [0.01, 1] should be reasonable for $\alpha$.
For $\gamma$, the performance is robust for $\gamma$ value
with $(0.001,1]$, but degrades when $\gamma$ becomes large.
This suggests both adversarial terms work.
%
\subsection{Unsupervised Domain Adaptation}

In many real-world applications, some domains may not have any annotated data at all. 
Thus it is important to extend the MDTC systems to tackle this extreme case, multi-source unsupervised domain adaptation.
In this experimental setting, we have multiple labeled source domains and one unlabeled target domain which
only has unlabeled instances. 
Since our proposed dual adversarial training method can leverage unlabeled data across different domains 
for feature representation learning with dual adversarial mechanisms,
it is expected to be useful for multi-source unsupervised domain adaptation.
Since there is no labeled data in the target domain, we only used the shared feature as the inputs 
to two classifiers $\mathcal{C}_1$ and $\mathcal{C}_2$ in this domain, while the domain-specific features are set to be 0s.

We conducted unsupervised domain adaptation experiments on the Amazon review dataset. 
We adopt the same setting as \cite{chen2018multinomial}. 
For each experiment, three out of the four domains are used as the labeled source domains,
and the remaining one domain is used as the unlabeled target domain. 
The test results are evaluated in the target domain.
We compared our proposed DACL method with several domain adaptation methods, 
including: (1) A MLP trained on the source domains, which serves as a baseline.
(2) Two single-source domain adaptation methods, marginalized denoising autoencoder (mSDA) \cite{chen2012marginalized} and domain adversarial neural network (DANN) \cite{ganin2016domain}. 
When training mSDA and DANN, we combined the data in the multiple source domains into a single source domain. 
(3) The state-of-the-art multi-source domain adaptation methods, 
multi-source domain adaptation neural networks (MDAN(H) and MADN(S)) \cite{zhao2017multiple} and 
MAN (MAN-L2 and MAN-NLL) \cite{chen2018multinomial}.  
The comparison results are reported
in Table~\ref{table_ref3}. 
We can see the average results of most multi-source domain adaptation methods 
are better than that of the single domain adaptation methods.
The proposed model outperforms 
all the comparison methods in three out of the total four domains. 
In terms of average accuracy, the proposed method outperforms all the other methods. 
This suggests our method has good capacity in performing multi-source unsupervised domain adaptation.

\section{Conclusion}

In this paper, we proposed a novel dual adversarial co-learning model 
to leverage all available resources across different domains for multi-domain text classification. 
The model uses a shared feature extractor across different domains to induce domain-invariant features
and uses domain-specific feature extractors to capture domain-dependent information. 
Two adversarial learning mechanisms are incorporated to boost feature representation learning,
which enforce multi-domain alignment through domain discriminator
and improve labeled-unlabeled data alignment through maximal prediction discrepancy minimization.
Experimental results on two MDTC benchmark datasets show the proposed approach can improve the performance of MDTC effectively. 
We also extended the work into multi-source unsupervised domain adaptation experiments,
which again demonstrates impressive performance.

\bibliographystyle{aaai}  
\bibliography{paperbib}

\end{document}